\documentclass[default,iicol]{sn-jnl}

\jyear{2023}

\theoremstyle{thmstyleone}

\theoremstyle{thmstyletwo}

\theoremstyle{thmstylethree}

\usepackage{overpic}
\usepackage{color}
\usepackage{colortbl}
\usepackage{marvosym}  
\usepackage{mathtools}
            
\newcommand{\figref}[1]{Fig.~\ref{#1}}
\newcommand{\tabref}[1]{Table~\ref{#1}}
\newcommand{\secref}[1]{Sec.~\ref{#1}}

\def\ie{\textit{i.e.},\ }
\def\eg{\textit{e.g.},\ }
\def\etc{\textit{etc}}
\def\etal{\textit{et al.}}
\def\etc{\textit{etc}}

\definecolor{mygray}{gray}{.9}
\definecolor{myRed}{RGB}{219, 68, 55}
\definecolor{myGreen}{RGB}{15, 157, 88}
\definecolor{myBlue}{RGB}{66, 133, 244}
\definecolor{iblue}{rgb}{0.06, 0.75, 1.0}
\definecolor{myBlack}{RGB}{0, 0, 0}

\newcommand{\yc}[1]{{\textcolor{myBlack}{#1}}}

\newcommand{\tabincell}[2]{\begin{tabular}{@{}#1@{}}#2\end{tabular}}

\newcommand{\ourmethod}{{\fontfamily{ppl}\selectfont
Drag\&Drop}}
\def\supp{\textcolor{magenta}{\texttt{\#SuppMaterial}}}
\graphicspath{{./Imgs/}{../../figure/}}

\newcommand\blfootnote[1]{
    \begingroup
    \renewcommand\thefootnote{}\footnote{#1}
    \addtocounter{footnote}{-1}
    \endgroup
}

\begin{document}

\title[Drag\&Drop]{\textbf{Acquiring Weak Annotations for Tumor
Localization in Temporal and Volumetric Data}}

\author[1]{\fnm{Yu-Cheng} \sur{Chou}} 

\author[1]{\fnm{Bowen} \sur{Li}}

\author[2]{\fnm{Deng-Ping} \sur{Fan}} 

\author[1]{\fnm{Alan} \sur{Yuille}}

\author[1]{\fnm{Zongwei} \sur{Zhou}$\textsuperscript{\Letter}$}

\affil[1]{\orgdiv{Department of Computer Science}, \orgname{Johns Hopkins University}, \orgaddress{\city{Baltimore}, \country{USA}}}

\affil[2]{\orgdiv{Computer Vision Lab}, \orgname{ETH Zürich}, \orgaddress{\city{Zürich}, \country{Switzerland}}}

\affil[]{\tt Project Page: \href{https://github.com/johnson111788/Drag-Drop}{https://github.com/johnson111788/Drag-Drop}}

\abstract{
Creating large-scale and well-annotated datasets to train AI algorithms is crucial for automated tumor detection and localization. 
However, with limited resources, it is challenging to determine the best type of annotations when annotating massive amounts of unlabeled data. 
To address this issue, we focus on polyps in colonoscopy videos and pancreatic tumors in abdominal CT scans; both applications require significant effort and time for pixel-wise annotation due to the high dimensional nature of the data, involving either temporary or spatial dimensions. 
In this paper, we develop a new annotation strategy, termed \ourmethod, which simplifies the annotation process to drag and drop. 
This annotation strategy is more efficient, particularly for temporal and volumetric imaging, than other types of weak annotations, such as per-pixel, bounding boxes, scribbles, ellipses, and points. 
Furthermore, to exploit our \ourmethod\ annotations, we develop a novel weakly supervised learning method based on the watershed algorithm. 
Experimental results show that our method achieves better detection and localization performance than alternative weak annotations and, more importantly, achieves similar performance to that trained on detailed per-pixel annotations. 
Interestingly, we find that, with limited resources, allocating weak annotations from a diverse patient population can foster models more robust to unseen images than allocating per-pixel annotations for a small set of images. 
In summary, this research proposes an efficient annotation strategy for tumor detection and localization that is less accurate than per-pixel annotations but useful for creating large-scale datasets for screening tumors in various medical modalities.
}


\maketitle

\section{Introduction}
\blfootnote{$\textsuperscript{\Letter}$ Corresponding author: zzhou82@jh.edu.}
Tumor detection and localization are often approached as a semantic segmentation task known as \textit{detection by segmentation}. 
The hypothesis is that identifying and delineating tumor boundaries can improve the tumor detection rate~\cite{xia2022felix}. 
However, this idea might not apply to all medical scenarios, particularly for screening purposes, in which it is more critical to predict the approximate location and size of the tumors rather than focusing on the accurate segmentation of tumor boundaries. 
For instance, polyp detection only requires the identification of the polyp, which can then be removed during the colonoscopy procedure~\cite{winawer1993prevention,rex2017colorectal}. 
In such cases, accurate segmentation of the polyp's boundary may not be necessary. 
However, a majority of public datasets for polyp detection provide per-pixel annotation for every polyp~\cite{vazquez2017benchmark,misawa2021development,ma2021ldpolypvideo,smedsrud2021kvasir,ji2022video}, which is exceptionally time-consuming and costly. 
Similar issues arise in other medical scenarios that focus on tumor detection but allocate annotations at the pixel level~\cite{lee2017curated,porwal2018indian,qu2023annotating}. 
This stresses the potential wastage of resources when using the detection by segmentation strategy for creating large-scale annotated datasets for tumor detection. 
We posit that for certain detection tasks, high precision in boundary segmentation is not crucial, and therefore per-pixel annotations may not be necessary.

On the contrary, weak annotations are more cost-effective and require less time than per-pixel annotations. 
We hypothesize that weak annotations are more appropriate for tumor detection and localization than the detection by segmentation strategy. We justify this point from three perspectives.
\textbf{Firstly}, with a certain budget, per-pixel annotations inevitably sacrifice data diversity and population due to the high annotation cost. 
Weak annotations allow for greater diversity and thus improve the tumor detection rate in minority cases, such as age (Tables~\ref{tab:budget}--\ref{tab:population}).
\textbf{Secondly}, the formulation of tumor segmentation can generate numerous false positives. 
Pixel-wise annotated datasets, e.g., KiTS~\cite{heller2020international}, only provide images with tumors.
This can create a bias where AI algorithms learn to predict tumors in every unseen image (\tabref{tab:specificity}). 
\textbf{Thirdly}, per-pixel annotations require significant time and resources to perform. 
Specifically, per-pixel annotations for pancreatic tumors from 3D volumetric CT scans require four minutes per subject, whereas weak annotations only require an average of two seconds per subject (\tabref{tab:comparison_weak_annotations}). 
Similarly, for polyp detection, weak annotations are eight times faster to perform than per-pixel annotations (2s~vs.~16s). 

While per-pixel annotation is dauntingly expensive and time-consuming, it is still widely adopted to train and test AI algorithms for tumor detection and localization~\cite{isensee2021nnu,chen2023towards,liu2023clip}.
This paper designs a new weak annotation strategy for high-dimensional data, such as temporal and volumetric medical images, by exploiting contextual information across dimensions. 
We call this strategy ``\ourmethod'' because it involves clicking on the tumor and then dragging and dropping to provide the approximate radius of the tumor. 
This annotation strategy is sufficient to capture the size and location of each tumor without requiring precise boundary segmentation. 
To utilize \ourmethod\ annotations, we further develop a weakly supervised framework based on the classical watershed algorithm, and it is optimized using the approximate tumor size and location constraints provided by \ourmethod. 
%
%
Our weakly supervised framework significantly reduces the impact of noisy labels that commonly occur at tumor boundaries in per-pixel annotations.
We demonstrate in the experiments that training using the weak annotations by \ourmethod, AI algorithms can perform similarly to pixel-wise annotations in tumor detection and localization tasks. 
We also show the superiority of our \ourmethod\ annotations over the previous weak annotation strategies, such as scribbles, points, bounding boxes, and ellipses annotations, in terms of tumor detection and localization efficacy (\tableautorefname~\ref{tab:comparison_weak_annotations}).

\section{Related Works}
Many efforts have been developed to detect lesions automatically.
Specifically, deep convolutional neural networks are successfully applied to segment tumors in the brain~\cite{havaei2017brain,myronenko20193d,wang2021transbts,jiang2022swinbts}, lung~\cite{jin2018ct,fan2020inf,wang2017central}, pancreas~\cite{zhu2019multi,zhang2020robust,li2023early}, liver~\cite{almotairi2020liver,li2015automatic}, polyp~\cite{ji2021progressively,fan2020pranet,ji2023deep,ji2022video,wei2021shallow}, \etc.
However, most methods require densely-labeled high-quality annotations to train the sophisticated network and achieve promising performance.
To reduce the annotation cost, some solutions~\cite{papandreou2015weakly,hsu2019weakly,zeng2019multi,wang2017learning,zhang2020weakly,chu2021improving,cheng2022pointly,li2019weakly,li2022box2mask,tang2018regularized,tang2018normalized} have been dedicated to utilizing weak supervisions with the devised loss functions, network architectures, and learning strategies. 
%

With the least amount of labeling effort, Chen~\etal~\cite{chen2022c} utilized category annotations based on class activation mapping (CAM) with a category-causality chain and anatomy-causality chain to overcome the challenges of co-occurrence phenomenon and the unclear boundary of object foreground and background in medical images.

Despite the promising results, the efficacy of the information contained within the category labels is insufficient in precisely localizing lesions in medical images.
Zhang~\etal~\cite{zhang2020weakly} further \yc{used} scribble labels based on an auxiliary edge detection task to localize edges explicitly.
To encourage the segmentation predictions to be consistent under different perturbations for an input image, Liu~\etal~\cite{liu2022weakly} proposed an uncertainty-aware mean teacher that is incorporated into the scribble-based segmentation method.
Similarly, Zhang~\etal~\cite{zhang2022cyclemix} proposed CycleMix, a framework for scribble that used consistency losses to regulate the mixup strategy with a dedicated design of random occlusion, to perform increments and decrements of scribbles.

Nevertheless, the lack of direct generalizability of commonly employed loss functions in fully supervised contexts to their weakly supervised counterparts poses a challenge in devising a robust framework for effective object localization. 
Consequently, Lu~\etal~\cite{lu2020geometry} propose a multi-task loss function that takes into account the area of the geometric shape, the categorical cross-entropy, and the negative entropy.
In line with the parallel concept, Chu~\etal~\cite{chu2021improving} proposed to jointly train a lesion segmentation model and a lesion classifier in a multi-task learning fashion, where the supervision of the latter is obtained by clustering the RECIST measurements of the lesions.

Apart from the above weak annotations, the bounding box is often used in the interactive pipeline.
Tian~\etal~\cite{tian2021boxinst} replace the original pixel-wise mask loss with the proposed projection and pairwise affinity mask loss to minimize the discrepancy between the projections of the ground-truth box and the predicted mask and exploit the prior that proximal pixels with similar colors are very likely to have the same category label.
Li~\etal~\cite{li2022box2mask} then proposed to iteratively learn a series of level-set functions to obtain accurate segmentation mask predictions.
Beyond bounding boxes, point clicks are the other time-efficient annotation forms most commonly used in interactive segmentation scenarios.
Chen~\etal~\cite{cheng2022pointly} proposed Implicit PointRend, which tackles the unique challenges of point supervision with implicit mask representation.

By contrast, our \ourmethod\ annotation is the first weak annotating strategy that focuses on high-dimensional data, such as temporal and volumetric medical images.
Different from low-dimensional data, high-dimensional data \yc{requires} significant effort and time for pixel-wise annotation due to the high-dimensional nature of the data, involving either temporary or spatial dimensions.
By leveraging contextual information across dimensions, our \ourmethod\ annotation enables manual labeling based on a single 2D annotation in high-dimensional volumetric data, eliminating the need for annotating on a slice-by-slice basis.
To demonstrate its efficacy, we compare our \ourmethod\ annotations with per-pixel annotations as an upper-bound performance reference and all the aforementioned weak annotations.

\begin{figure*}[!t]
  \centering
  \includegraphics[width=\linewidth]{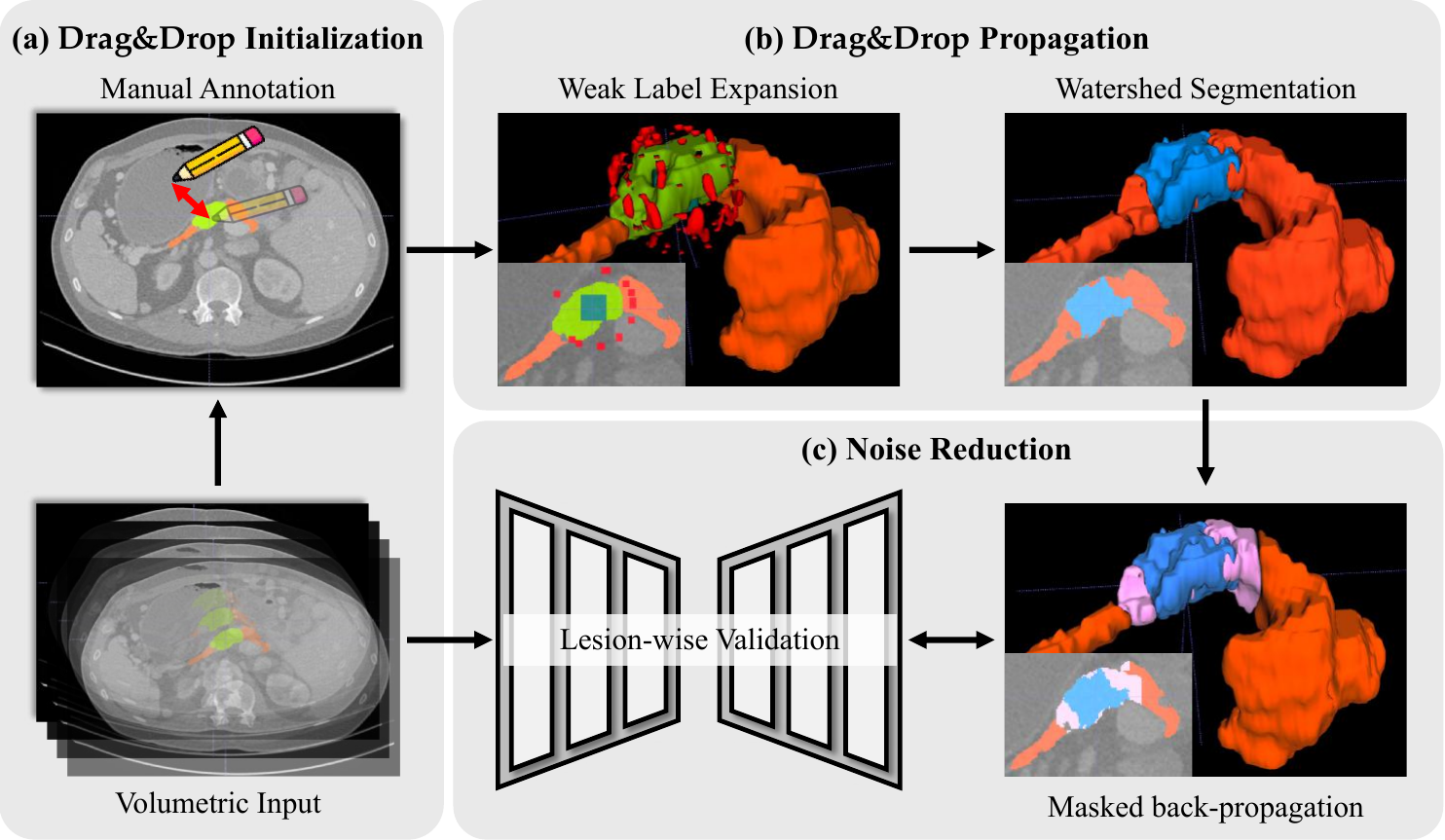}
  \caption{
    Annotation and propagation process of the proposed \ourmethod.
    The pancreas and tumor are represented by the colors orange and green, respectively.
    Given the weak label, we expand it to a lesion marker (teal green) and background markers (red).
    We then utilize the marker-based watershed algorithm to generate the initial segmentation area (blue), following which dilated tumors (pink) are applied to compute the masked back-propagation.
  }\label{fig:method}
\end{figure*}

\section{Method}
\label{sec:method}

To reduce manual annotation efforts, we propose a novel annotation strategy termed \ourmethod\ (\secref{sec:dd}) and a weakly supervised learning framework, consisting of 3D annotation propagation (\secref{sec:3d_anno_prop}) and noise reduction (\secref{sec:noise_red}) to achieve a better cost-accuracy trade-off.
%

\subsection{Drag\&Drop Initialization---\\Manual Annotation}\label{sec:dd}
According to empirical evidence, the annotating process can be segmented into two distinct stages: \textit{edge determination} and \textit{manual labeling}. 
When dealing with high-dimensional data, the annotator first determines the edge of a lesion and manually draws the labels according to the annotating methods.
Weak annotating methods spanning multiple dimensions can substantially reduce the \yc{number} of labels and shorten the manual labeling time compared to the lower-dimensional ones that are based on a slice-by-slice basis.
As a result, to facilitate manual labeling, we instructed annotators \textit{only} to provide a single 2D annotation in high-dimensional volumetric data and propagate it by exploiting 3D contextual information across dimensions (more details in~\secref{sec:3d_anno_prop}).
Specifically, we ask the annotator to \fontfamily{ppl}{\selectfont{drag}} a radius from the central point of the lesion and \fontfamily{ppl}{\selectfont{drop}} it at the boundary.
\figref{fig:method}-(a) depicts the detailed \ourmethod\ annotation process.
Compared to other strategies that need multiple 2D or 3D annotations, our \ourmethod\  only requires one 2D label per lesion within the high-dimensional data, resulting in a significant reduction in manual labeling time.

Considering edge determination time, although Roth~\etal~\cite{roth2021going} attempted to use extreme points in high-dimensional data, it still necessitates six click locations lying on the surface of the target, demanding annotators precisely distinguish the lesion area.
Similarly, Zhou~\etal~\cite{zhou2019integrating} tried to simplify the video annotation process with six clicks on the lumen-intima and media-adventitia interfaces, which also makes a high degree of precision in the observation of the video frame necessary.
In contrast, our \ourmethod\ only requires annotators to provide an approximate radius from the central point without the need for careful observation of the lesion's boundaries.
As a result, our method significantly reduces the time required for edge determination.
In this study, we automatically extract the central location and its corresponding radius from a given ground truth mask for training.
In order to simulate user interaction, we add some Gaussian noise to $x$, $y$, and $z$ directions.

\yc{To ensure the quality of \ourmethod, the annotators must adhere to the following guidance for accurately labeling lesions.
1. First, the annotators should locate the lesions and screen through multiple scans to estimate their size and three-dimensional center.
2. Next, as \ourmethod is adaptable to multiple classes and targets, the annotators can conveniently annotate lesions in arbitrary dimensions.
Specifically, the given radius should encompass the lesion in all dimensions, and multi-class annotations can be applied accordingly when multiple categories of lesions are present within an image.
3. Last, the annotator could refine the annotation by adding and removing the annotation mask \yc{by} simply clicking the foreground and background, respectively.
By following the aforementioned steps, the introduction of size discrepancies by different annotators can be minimized or eliminated, thereby preserving the quality of weak annotations when utilizing \ourmethod.
}

In terms of other weak annotations, we follow the same method in Ji~\etal~\cite{ji2022video} to generate the annotations automatically.
Specifically, the bounding box is generated by calculating the lower-/upper-bound of the object mask.
Following Cheng~\etal~\cite{cheng2022pointly}, $10$ points are randomly generated within the area of the bounding box and classified by the object mask to collect the positive and negative points.
Next, we apply LIN algorithm~\cite{fitzgibbon1996buyer} to calculate the ellipse that fits (in a least-squares sense) a set of 2D points best of all.
As for scribble, we use two high-degree curves to indicate the foreground and background, respectively. 
To ensure the objectivity of various annotators, we adopt linear or quadratic functions to randomly create the above curves in the positive/negative region.

\begin{table*}[t!]
  \centering
  \scriptsize
  \renewcommand{\arraystretch}{1.2}
  \setlength\tabcolsep{8.3pt}
  \caption{Given a certain annotation budget, our \ourmethod\ strategy outperforms the per-pixel annotation by a large margin in tumor detection and localization. More importantly, compared with per-pixel annotations, \ourmethod\ improves tumor segmentation from 0.43 to 0.54 for pancreatic tumor detection measured by DSC scores.
    }
  \begin{tabular}{c|c||cccc|cccc} 
  \toprule
  & 
  & \multicolumn{4}{c|}{\tabincell{c}{Lesion-level}}  
  & \multicolumn{4}{c}{\tabincell{c}{Patient-level}} \\
  \hline
  Method & Strategy
  & Sen. & Spe. & Pre. & F1 
  & Sen. & Spe. & Pre. & F1  \\
  \hline
  \multicolumn{10}{c}{\tabincell{c}{\textbf{JHH}}~\cite{xia2022felix}} \\
  \hline
  nnUNet & Per-pixel
  &0.611 &0.339 &0.422 &0.522 &0.765 &\textbf{0.702} &0.752 &0.613 \\
  \rowcolor{mygray}  
  \textbf{nnU-Net} & \textbf{\ourmethod}
  &\textbf{0.715} &\textbf{0.429} &\textbf{0.575} &\textbf{0.610} &\textbf{0.886} &0.649 &\textbf{0.749} &\textbf{0.735} \\
  
  \hline
  \multicolumn{10}{c}{\tabincell{c}{\textbf{SUN-SEG}~\cite{ji2022video}}} \\
  \hline
  PNS+ & Per-pixel
  &0.681 &0.589 &0.434 &0.546 &0.759 &0.698 &0.543 &0.621 \\
  \rowcolor{mygray}
  \textbf{PNS+}& \textbf{\ourmethod}
  &\textbf{0.719} &\textbf{0.677} &\textbf{0.512} &\textbf{0.595} &\textbf{0.804} &\textbf{0.790} &\textbf{0.644} &\textbf{0.668} \\
    \bottomrule
  \end{tabular}
  \label{tab:budget}
\end{table*}

\subsection{Drag\&Drop Propagation---\\Watershed Algorithm}\label{sec:3d_anno_prop}
To propagate the \ourmethod\ annotations to pseudo labels, we adopt a marker-based watershed transformation algorithm~\cite{meyer1994topographic} to separate the image into positive and negative regions.
Compared to other segmentation methods, the watershed algorithm does not require parameter tuning and can use markers as a form of user guide to refine the segmentation boundaries.
That is, we can use the given \ourmethod\ annotation to generate pseudo labels with a relatively precise location.
The watershed algorithm views an image as a topographic landscape with ridges and valleys. 
%
%
The marker-based watershed algorithm utilizes this hierarchical representation to decompose an image into catchment basins by flooding an image from the markers until it reaches the boundaries, \ie watershed line, of the regions. 
In specific, given an input $I$, lesion markers $m^l_i$, and background markers $m^b_j$, the watershed line can be defined as the set of points of the support of $I$ \yc{that} do not belong to any catchment basin:
\begin{equation}
\begin{multlined}
Wsh(I, m^l_i, m^b_j)=\\supp(I) \cap\left[\bigcup_i\left(CB\left(m^l_i\right)\right)\bigcup_j\left(CB\left(m^b_j\right)\right)\right],
\end{multlined}
\end{equation}
where $CB(m)$ and $supp(I)$ stand for the catchment basin of the marker $m$ and support of $I$, respectively.

To accurately segment the boundary of the lesions, we dilate the \ourmethod\ annotations into lesion and background markers in the 3D space as the initial flooding points.
Given a central point and radius of a lesion in a high-dimensional input, we first generate a 3D sphere and sample set the surface of the sphere as the background markers $m^b_j$.
Specifically, we set $j$ as the number of integer points on the surface and randomly sample the background markers.
Then we adaptively dilate the central point according to a certain ratio $N$ of the given radius to avoid under-segmentation and create the lesion marker $m^l_i$, ensuring that the result adequately co
vers the region of interest.
To remove noise while preserving the boundary information, we further applied a morphological gradient to the input $I$ before watershed segmentation.
By utilizing the lesion and background markers for each lesion in conjunction with the denoised input, we recursively applied a marker-based watershed algorithm on each lesion to propagate the markers to pseudo labels.
Segment process and results are illustrated in~\figref{fig:method}-(b).

\begin{table*}[t!]
  \centering
  \scriptsize
  \renewcommand{\arraystretch}{1.25}
  \setlength\tabcolsep{4.6pt}
  \caption{Lesion-level performance comparison regarding to the age distribution on the JHH dataset. 
  Our \ourmethod\ enables more diversity and thereby enhances the detection rate in minority cases (highlighted in \textbf{bold}).
  We used nnU-Net~\cite{isensee2021nnu} for per-pixel annotations.
  Note that in the age range of 45, the test result is 0 because all the cases were healthy and had no tumors.
    }
    
  \begin{tabular}{c|c||cccccccccc} 
  \toprule
  & Age 
  & [35 40)
  & [40 45) & [45 50) & [50 55) & [55 60)
  & [60 65) & [65 70) & [70 75) & [75 80)
  & [80 85)\\
  \hline
  & Num.
  & 20  & 20  & 27  & 21  & 25 & 16 & 17 & \textbf{6} & \textbf{12} & \textbf{6} \\
  \hline
  \multirow{2}{*}{Pre.} &
  Per-pixel
  &0.166 &0.095 &0.000 &0.307 &0.200 &0.600 &0.562 &0.499 &0.428 &0.454\\
  & \ourmethod
  &0.125 &0.133 &0.000 &0.333 &0.333 &0.368 &0.705 &\textbf{1.00} &\textbf{0.611} &\textbf{0.600} \\
  \hline
  \multirow{2}{*}{F1} &
  Per-pixel 
  & 0.286& 0.174& 0.000& 0.459& 0.316& 0.666& 0.624 & 0.666 & 0.558 & 0.594 \\
  & \ourmethod
  & 0.222  & 0.235 & 0.000  & 0.484  & 0.429 & 0.494 & 0.828 & \textbf{1.00} & \textbf{0.720}& \textbf{0.750}\\
    \bottomrule
  \end{tabular}
  \label{tab:population}
\end{table*}

\begin{table*}[t!]
  \centering
  \scriptsize
  \renewcommand{\arraystretch}{1.15}
  \setlength\tabcolsep{9.0pt}
  \caption{ 
  Training data distribution in terms of positive and negative samples.
  The methods trained on both positive and negative data can generally surpass the one \textit{only} trained on positive data.
    }
  \begin{tabular}{c|cc||cccc|cccc} 
  \toprule
  & \multicolumn{2}{c||}{\tabincell{c}{Distribution}}  
  & \multicolumn{4}{c|}{\tabincell{c}{Lesion-level}}  
  & \multicolumn{4}{c}{\tabincell{c}{Patient-level}} \\
  \hline
  Strategy & Pos. & Neg.
  & Sen. & Spe. & Pre. & F1 
  & Sen. & Spe. & Pre. & F1  \\
  \hline
  \multicolumn{11}{c}{\tabincell{c}{\textbf{JHH}~\cite{xia2022felix}}} \\
  \hline
  \multirow{2}{*}{Per-pixel} & $1168$ & $0$
  &0.789 &0.475 &0.622 &0.656 &0.920 &0.709 &0.789 &0.771 \\
  & $1168$ & $515$ 
  &0.789 &0.729 &0.780 &0.620 &0.937 &0.980 &0.982 &0.351 \\
  
  \hline
  \multirow{2}{*}{\ourmethod} & $1168$ & $0$
  &0.715 &0.429 &0.575 &0.610 &0.886 &0.649 &0.749 &0.735 \\
  & $1168$ & $515$  
  &0.769 &0.691 &0.747 &0.620 &0.914 &0.953 &0.958 &0.478 \\
  
  \hline
  \multicolumn{11}{c}{\tabincell{c}{\textbf{SUN-SEG}~\cite{ji2022video}}} \\
  \hline
  
  \multirow{2}{*}{Per-pixel} & $189$ & $0$
  &0.783 &0.681 &0.551 &0.634 &0.869 &0.748 &0.620 &0.700\\
  
  & $189$ & $318$
  &0.763 &0.813 &0.669 &0.640 &0.846 &0.900 &0.801 &0.671  \\
  
  \hline
  
  \multirow{2}{*}{\ourmethod} & $189$ & $0$
  &0.719 &0.677 &0.512 &0.595 &0.804 &0.790 &0.644 &0.668 \\
  
  & $189$ & $318$
  &0.716 &0.701 &0.530 &0.600 &0.804 &0.819 &0.677 &0.669 \\
  
    \bottomrule
  \end{tabular}
  \label{tab:specificity}
\end{table*}

\subsection{Noise Reduction in Training and Evaluation}\label{sec:noise_red}
Even though the watershed algorithm can detect the boundaries of lesions, it remains vulnerable to noise in the image, particularly in cases of individual variations in the approximate radius drawn by the annotators, which is pronounced given their irregular shapes.
%
To this end, we propose a masked back-propagation that computes the gradient only on the lesion and background regions that are confidently provided by the watershed algorithm; no gradient on the uncertain regions. 
In detail, we dilate pseudo labels with $M$ kernel size and ignore the gradient of the dilated area (which is the uncertain region) during back-propagation.
Due to the constraint provided by background markers on the sphere, the pseudo label displayed a lower rate of false positives. As a result, we chose not to perform erosion to obtain the gradient-free region.
This strategy considerably increases the robustness of the proposed method and refines the segmentation result of the watershed algorithm by largely reducing the number of false negatives (Sensitivity: $0.98$, Specificity: $1.00$; detailed in~\tabref{tab:ablation_study}).

In addition, due to the absence of per-pixel labels, the performance validation of the model during training is susceptible to biases and inaccuracies, leading to the misestimation of the model's performance and the erroneously selected parameters.
To address this issue comprehensively and obtain optimal detection performance, we propose to adopt a lesion-wise metric for validation purposes.
Compared to the commonly adopted pixel-wise metric~\cite{chu2021improving}, lesion-wise one has a high tolerance for pixel-wise imprecision in the pseudo label and thus poses the capability to achieve the best model performance during training.

\section{Experiment \& Result}
\label{sec:experiment_result}

\subsection{Benchmarking}

\noindent\textbf{\textit{Dataset.}}
We adopt two large-scale high-dimensional datasets in our experiments, including a private 3D volumetric CT dataset (\ie JHH~\cite{xia2022felix}) and a colonoscopy video dataset (\ie SUN-SEG~\cite{ji2022video}).
JHH dataset includes $2,426$ CTs of $1,213$ patients with per-pixel annotation of pancreatic tumor, containing classes of pancreas, PDAC, and Cyst.
%
%
Pancreatic protocol CTs were retrospectively identified from clinical, pathological, and radiological databases compiled between $2003$ and $2020$.
Patients with pancreatic tumors between $2011$ to $2020$ were scanned with a dual-source MDCT scanner (Somatom Definition, Definition Flash, or Force, Siemens Healthineers), and the patients between $2003$ to $2010$ were scanned on a 16- or 64-slice MDCT scanner (Somatom Sensation 16 or 64, Siemens Healthineers).
Both arterial and venous phase images were collected for patients, resulting in $2,426$ abdominal CT scans. 
Arterial phase imaging was performed with fixed delay or bolus triggering, usually between $25 \sim 30$ seconds post-injection, and venous phase imaging was performed at $60$ seconds.
Each CT scan consists of $319 \sim 1,051$ slices of $512 \times 512$ pixels, and have a voxel spatial resolution of $([0.523 \sim 0.977] \times [0.523 \sim 0.977] \times 0.5)$mm$^3$
We randomly split $1683$, $420$, and $323$ cases for training, validation, and testing, respectively.
It is noteworthy that for the sake of simplicity, the combined class of PDAC and Cyst is reported during the evaluation process.

\begin{figure*}[t!]
  \centering
  \includegraphics[width=0.97\linewidth]{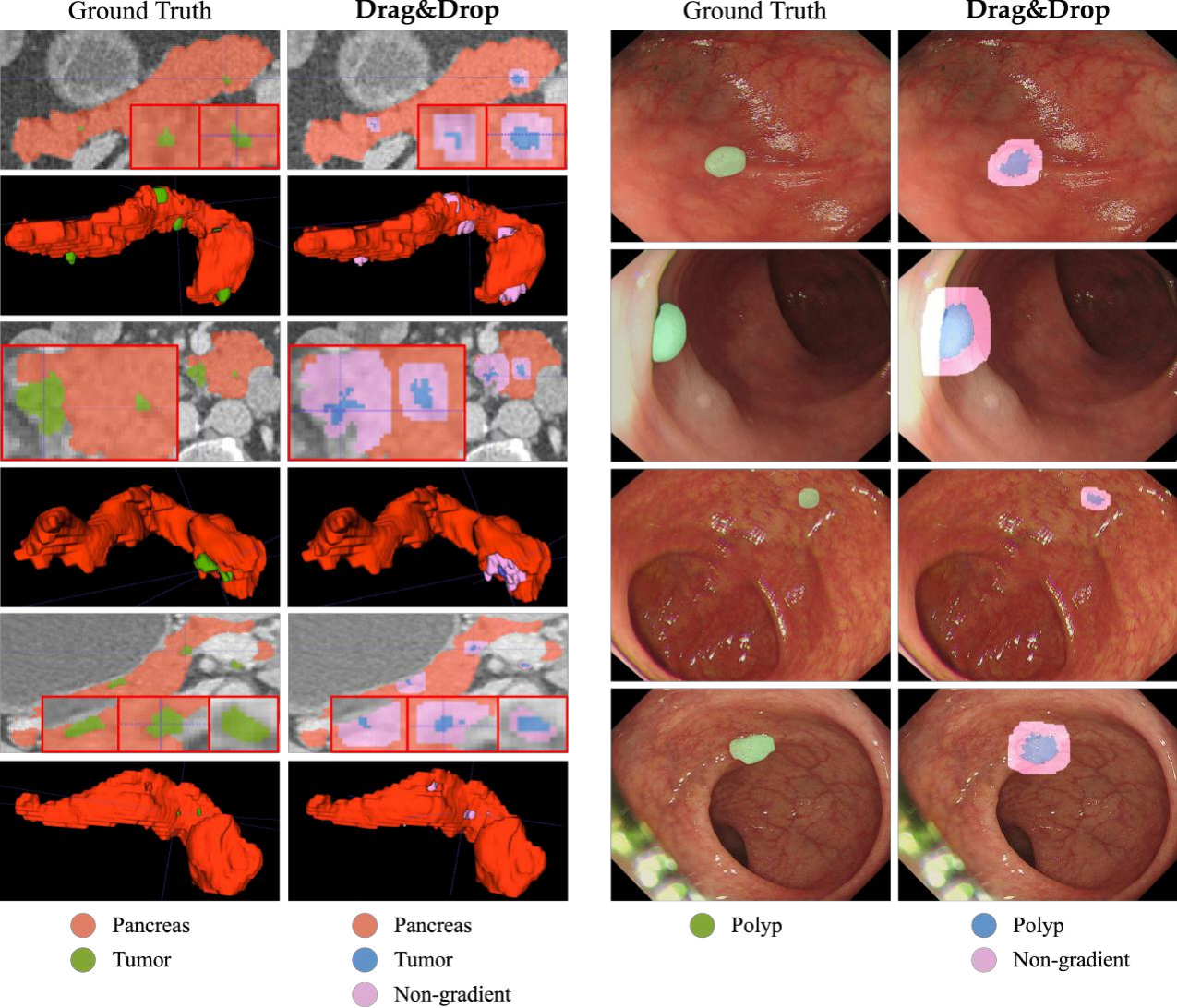}
  \caption{
  Illustration of \ourmethod\ annotations on the JHH and SUN-SEG datasets.  
  %
  %
  }\label{fig:qual}
\end{figure*}

SUN-SEG dataset, on the other hand, offers $1,106$ video clips with $158,690$ frames total.
The colonoscopy videos in SUN-SEG dataset are from Showa University and Nagoya University database, the largest video polyp dataset for the detection task.
The videos are captured by the high-definition endoscope (CF-HQ290ZI \& CF-H290ECI, Olympus) and video recorder (IMH-10, Olympus), providing videos of various polyp sizes at dynamic scenarios, such as imaging at different focusing distances and speeds.
SUN-SEG dataset contains annotations of attributes, object masks, bounding boxes, boundaries, scribbles, and polygons.
According to the origin bounding box labels of the SUN-database~\cite{misawa2021development}, ten experienced annotators are instructed to offer various labels using Adobe Photoshop.
Then, three colonoscopy-related researchers re-verify the quality and correctness of these initial annotations.
%
%
We follow the same data splitting protocol in~\cite{ji2022video} and re-split the videos in a finer granularity, resulting in $507$ and $126$ cases respectively for training and validation.
As for the testing set, we merge the four original testing sets and have $1,050$ cases.

\begin{table*}[t!]
  \centering
  \scriptsize
  \renewcommand{\arraystretch}{1.15}
  \setlength\tabcolsep{7.5pt}
  \caption{We also present the cost-accuracy trade-off.
  Note that the baselines adopt low-dimensional annotation strategies and are designed specifically for 2D input, resulting in severe performance degradation on the JHH dataset.
  In contrast, nnUNet is highly adaptable to 3D volumetric data and thus \yc{achieves} satisfying results.
    }
  \begin{tabular}{r|ll||cccc|cccc} 
  \toprule
  & & 
  & \multicolumn{4}{c|}{\tabincell{c}{Lesion-level}}  
  & \multicolumn{4}{c}{\tabincell{c}{Patient-level}} \\
  \hline
  Method & Strategy & Time 
  & Sen. & Spe. & Pre. & F1 
  & Sen. & Spe. & Pre. & F1  \\
  \hline
  \multicolumn{11}{c}{\tabincell{c}{\textbf{JHH}}} \\
  \hline
  SODSA & Scribble & $0'38''$ 
  & 0.086 & 0.005 & 0.025 & 0.048 & 0.121 & 0.027 & 0.126 & 0.200 \\
  WeakMTL & Ellipse & $0'46''$ 
  &0.054 &0.035 &0.010 &0.020 &0.069 &0.313 &0.104 &0.165 \\
  PointSup & 10 Points  & $1'17''$ 
  &0.525 &0.411 &0.562 &0.507 &0.514 &0.460 &0.524 &0.507\\
  Box2Mask & B-Box & $2'06''$ 
  &0.488 &0.384 &0.414 &0.481 &0.659 &0.693 &0.712 &0.543\\
  nnUNet & Per-pixel & $4'15''$  
  &0.789 &0.475 &0.622 &0.656 &0.920 &0.709 &0.789 &0.771\\
  \textbf{Ours} & \textbf{\ourmethod} & \textbf{$\mathbf{0'02''}$} 
  &0.715 &0.429 &0.575 &0.610 &0.886 &0.649 &0.749 &0.735 \\
  \hline
  \multicolumn{11}{c}{\tabincell{c}{\textbf{SUN-SEG}}} \\
  \hline 
  SODSA & Scribble & $0'03''$  
  &0.701 &0.647 &0.496 &0.582 &0.783 &0.718 &0.567 &0.640\\
  WeakMTL & Ellipse & $0'04''$
  &0.709 &0.485 &0.389 &0.522 &0.792 &0.576 &0.469 &0.595\\
  PointSup & 10 Points & $0'07''$
  &0.712 &0.676 &0.520 &0.595 &0.795 &0.754 &0.604 &0.656\\
  Box2Mask & B-Box & $0'09''$  
  &0.684 &0.600 &0.422 &0.539 &0.772 &0.771 &0.615 &0.645\\
  PNS+ & Per-pixel & $0'16''$  
  &0.783 &0.681 &0.551 &0.634 &0.869 &0.748 &0.620 &0.700\\
  \textbf{Ours} & \textbf{\ourmethod} & \textbf{$\mathbf{0'02''}$} 
  &0.719 &0.677 &0.512 &0.595 &0.804 &0.790 &0.644 &0.668 \\
  \bottomrule
  \end{tabular}
  \label{tab:comparison_weak_annotations}
\end{table*}

\smallskip\noindent\textbf{\textit{Evaluation metrics.}} 
The assessment of the detection performance involves lesion-level and patient-level evaluation, which include
(a) \textbf{Sensitivity} (Sen.) to evaluate the true positive prediction $TP$ of overall lesion areas and penalize the false negative prediction $FN$ to increase the screening ability of the algorithm, which is defined as:
$$\rm{Sen.} = \frac{ TP }{TP + FN}.$$ 
(b) \textbf{Specificity} (Spe.) to calculate the ratio of actual negatives accurately diagnosed $TN$ and penalize the false positive $FP$ to suppress the overdiagnosis.
It can be defined as:
$$\rm{Spe.} = \frac{ TN }{TN + FP}.$$ 
(c) \textbf{Precision} (Pre.). 
Different from sensitivity evaluates the true positive prediction of overall lesion areas, precision focuses on the ratio of the true positive prediction in overall prediction areas. 
Precision can be seen as a measure of quality and sensitivity as a measure of quantity. 
Higher precision means that an algorithm returns more relevant results than irrelevant ones, and high sensitivity means that an algorithm returns most of the relevant results (whether or not irrelevant ones are also returned).
It can be defined as:
$$\rm{Pre.} = \frac{ TP }{ TP + FP }.$$
(d) \textbf{F1-score} (F1). 
F1-score is an alternative evaluation metric that assesses the predictive skill of an algorithm by elaborating on its class-wise performance rather than an overall performance as done by accuracy.
It is a popular metric to use for classification models as it provides robust results for both balanced and imbalanced datasets, unlike accuracy.
It can be defined as:
$$\rm{F1} = \frac{ 2 \times TP}{2 \times TP + FN + FP}.$$

In addition, we adopt Free-Response Receiver Operating Characteristic (FROC) analysis, an evaluation approach balancing both sensitivity and false positives.
The FROC analysis is reported with sensitivities at various false positive levels. 
We also adopt the age information provided in the JHH dataset to compare the model performance in different population distributions.
%

\smallskip\noindent\textbf{\textit{Implementation and baselines.}} 
For pancreatic tumor detection and polyp detection, we adopt the state-of-the-art per-pixel segmentation methods (\ie\ nnUNet~\cite{isensee2021nnu} and PNS+~\cite{ji2022video}, respectively) as our model to validate the efficacy of the proposed annotation strategy.
We follow the default training setting with the standard binary cross-entropy loss and the Dice loss in the learning process.
We set $N=0.2$ and $M=0.5$ after extensive ablation studies (see~\tabref{tab:ablation_study}).
As previous studies have demonstrated outstanding outcomes in pancreas segmentation, we utilize a masked cropped volume to assist in the detection of pancreatic tumors.
And in polyp detection, we apply \ourmethod\ at the frame level.
We re-train five weakly supervised baselines with the same data used by our \ourmethod, under their default settings, for fair comparison.

\begin{table*}[h]
  \centering
  \scriptsize
  \renewcommand{\arraystretch}{1}
  \setlength\tabcolsep{19.2pt}
  \caption{
  Ablation studies. $N$ and $M$ stand for the ratio of dilation kernel size to the sphere radius for a central point and for pseudo labels, respectively.
    }
  \begin{tabular}{r|ll||cccc} 
  \toprule
  & \multicolumn{2}{c||}{\tabincell{c}{Variant}}  
  & \multicolumn{4}{c}{\tabincell{c}{Pixel-level}}  
  \\
  \hline
  No. & $N$ & $M$ 
  & Dice & Precision & Sensitivity & Specificity \\
  \hline
  \multicolumn{7}{c}{\tabincell{c}{\textbf{JHH}}} \\
  \hline

  
  
  \#1 & \cellcolor{iblue!15}$0.1$ & $0.5$
  & 0.677 & 0.688 & 0.847 & 0.997 \\   
  \#2 & \cellcolor{iblue!15}$0.2$ & $0.5$
  & 0.792 & 0.737 & 0.979 & 0.997 \\
  \#3 & \cellcolor{iblue!15}$0.4$ & $0.5$
  & 0.790 & 0.694 & 0.998 & 0.995 \\

  \hline
  \#4 & $0.2$ & \cellcolor{iblue!15}$0.4$
  & 0.734 & 0.737 & 0.931 & 0.997 \\
  \#5 & $0.2$ & \cellcolor{iblue!15}$0.5$
  & 0.792 & 0.737 & 0.979 & 0.997 \\
  \#6 & $0.2$ & \cellcolor{iblue!15}$0.6$
  & 0.824 & 0.737 & 0.992 & 0.997 \\
  
  \hline

  \multicolumn{7}{c}{\tabincell{c}{\textbf{SUN-SEG}}} \\
  \hline
  \#7 & \cellcolor{iblue!15}$0.1$ & $0.5$
  &0.784 &0.787 &0.893 &0.996 \\
  \#8 & \cellcolor{iblue!15}$0.2$ & $0.5$
  &0.845 &0.759 &0.981 &0.994 \\
  \#9 & \cellcolor{iblue!15}$0.4$ & $0.5$
  &0.776 &0.664 &0.999 &0.988 \\
  \hline
  \#10 & $0.2$ & \cellcolor{iblue!15}$0.4$
  &0.820 &0.763 &0.936 &0.995 \\
  \#11 & $0.2$ & \cellcolor{iblue!15}$0.5$
  &0.845 &0.759 &0.981 &0.994 \\
  \#12 & $0.2$ & \cellcolor{iblue!15}$0.6$
  &0.853 &0.759 &0.996 &0.994 \\
  
  
  \bottomrule
  \end{tabular}
  \label{tab:ablation_study}
\end{table*}

\subsection{Population Diversity}
To evaluate the performance gain provided by data diversity in our \ourmethod\ strategy, we calculate the number of per-pixel labels that can be annotated in the same amount of time.
The annotation time can be defined as detection time plus the manual labeling time.
Since pancreatic tumors are more camouflaged and require more time to be detected, we set the detection time as $30''$ and $2''$ for pancreatic tumors and polyps, respectively.
With the manual labeling time of $4'15''$ and $2''$, we can have the number of per-pixel labeled and \ourmethod\ labeled cases for pancreatic tumor ($131$ vs. $1,168$) and polyp detection ($33$ vs. $189$).
The results in~\tabref{tab:budget} and~\tabref{tab:population} both illustrate that the utilization of weak annotations can promote greater diversity and thus enhance results.

\subsection{Bias Mitigation}
We further discuss the bias \yc{from} the data distribution of positive and negative samples.
We compare the model trained with only positive data and the one trained with both positive and negative data.
As shown in~\tabref{tab:specificity}, the methods trained on both positive and negative data can generally surpass the one \textit{only} trained on positive data.
Specifically, the results \yc{have} increased significantly, especially for specificity (0.649 vs. 0.958), suggesting that the well-adopted `detection by segmentation'  strategy is \yc{unsuitable} for tumor detection on pixel-wise annotated datasets that only contain images with tumors.




\begin{figure*}[t!]
  \centering
  \includegraphics[width=0.8\linewidth]{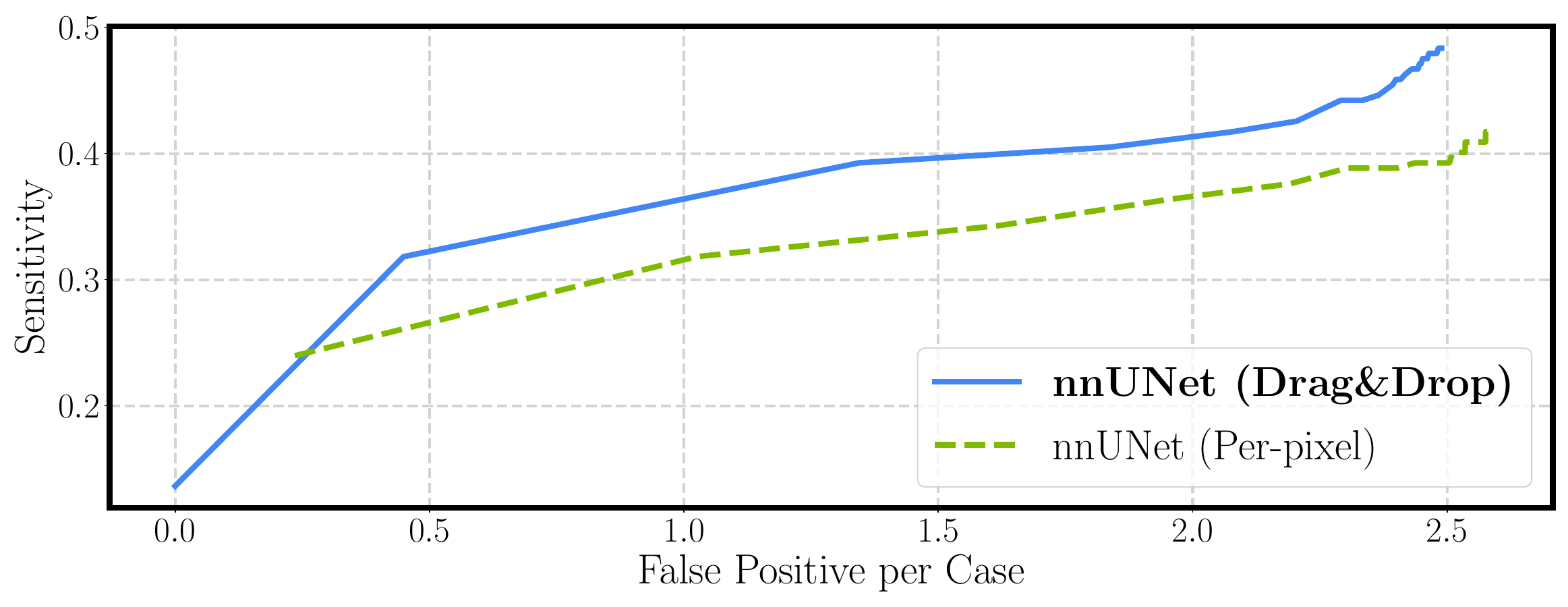}
  \caption{
  FROC curves of nnUNet trained by \ourmethod\ and per-pixel annotation strategy with a certain budget on the JHH dataset.
  }\label{fig:FROC}
\end{figure*}

\subsection{Cost-accuracy Trade-off}
%
Last, we compare the proposed \ourmethod\ to other weak annotations in terms of the cost-accuracy trade-off.
As shown in~\tabref{tab:comparison_weak_annotations}, our \ourmethod\ offers a distinct advantage over other strategies. 
Requiring only $2$ seconds of annotation, our method achieves similar performance to all the other weak annotation strategies that demand more annotation time.
Therefore, our method is well-suited for large-scale projects where cost-effectiveness and efficient use of resources are critical considerations.
In terms of the annotation time, two medical researchers are instructed to offer various annotations and calculate the annotation time.
Each annotation type is conducted on three randomly selected cases and presents the average annotation time.

\subsection{Detection Error Trade-off}
As shown in~\figref{fig:FROC}, Free-response ROC (FROC) of nnUNet trained by \ourmethod\ and per-pixel annotation strategy with a certain budget on JHH dataset (see \tabref{tab:budget}).
The result indicates that the proposed \ourmethod\ annotation strategy generally outperforms the per-pixel annotation at most threshold settings.
Instead of a fixed threshold, the threshold range is determined by the confidence scores of local maximum predictions, with the lowest and highest scores defining the range.

\subsection{Ablation Study}
To investigate the optimal setting of the hyperparameters, we conduct the ablation study of the ratio of dilation kernel size to the sphere radius for a central point $N$ and pseudo labels $M$. 
As shown in~\tabref{tab:ablation_study}, we first ablate the ratio of dilation kernel size to the sphere radius for a central point $N$.
We observe that \#2 and \#8 outperform other settings (\ie \#1, \#3, and \#7, \#9, respectively) in terms of dice metric, which is our main focus.
Therefore, we set $N=0.2$ to obtain the best performance.
We further investigate the setting of $M$. 
With the increase of dilation kernel size, false negative decreases, but true negative and supervision signal also decrease owing to the increase of masked area, which hampers the learning process.
As a result, we set $M=0.5$ for the balance of noise reduction and learning efficiency.

\subsection{Discussion}
\begin{figure}[t!]
  \centering
  \includegraphics[width=\linewidth]{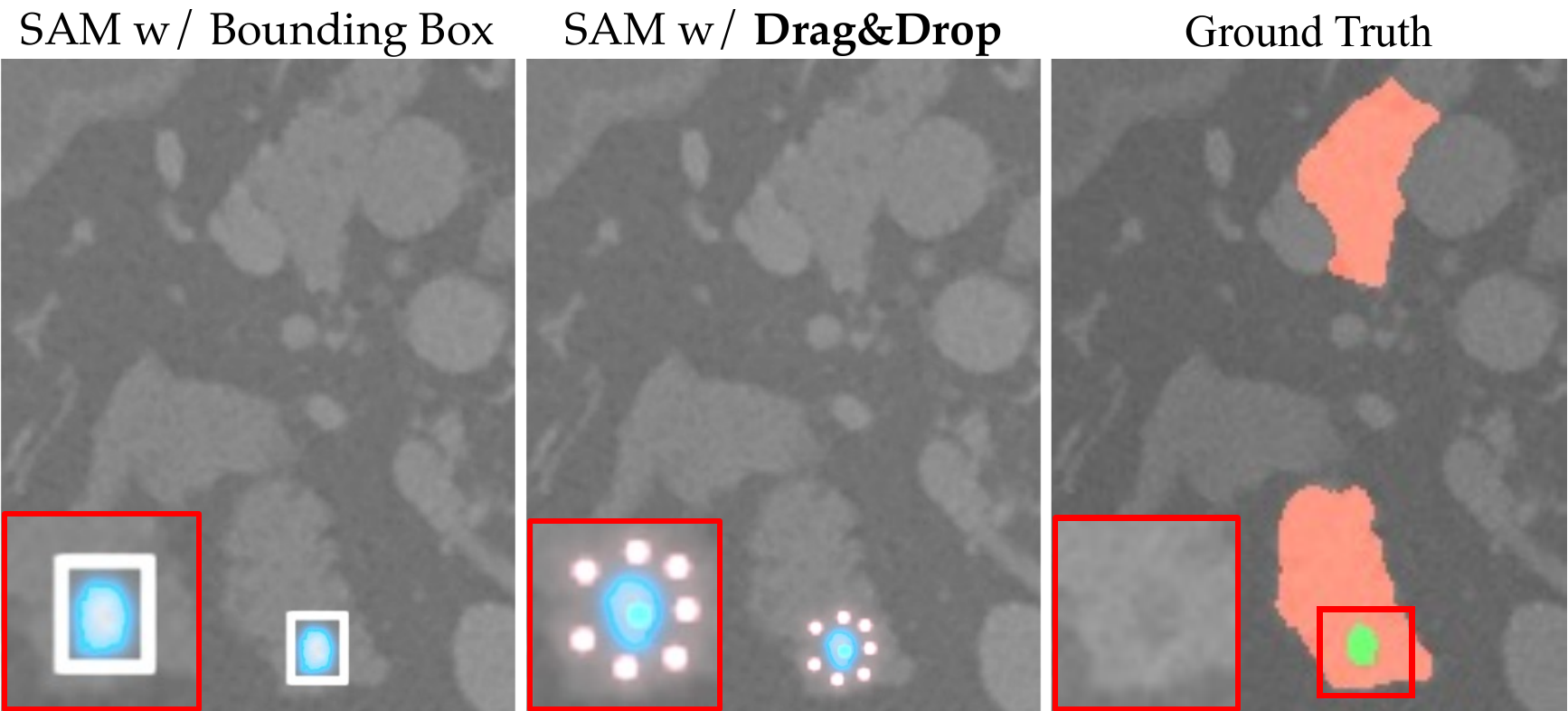}
  \caption{
  \yc{Illustration of the segmentation result of SAM using \ourmethod\ and bounding box as a prompt.}  
  }\label{fig:rebuttal}
\end{figure}

\yc{
To conduct a more comprehensive comparison with other strategies, we discuss the strengths and weaknesses based on the state-of-the-art segmentation methods, SAM~\cite{kirillov2023segment} and MedSAM~\cite{ma2023segment}.
SAM and MedSAM are 2D segmentation methods where users can use a prompt to specify the segmentation target.
Namely, the proposed \ourmethod\ annotation strategy could serve as a prompt to enhance flexibility and adaptability.
Compared to using bounding boxes or points as prompt, the proposed \ourmethod\ is more efficient, particularly for temporal and volumetric imaging.
Besides, SAM does not impose semantic constraints for labeling objects, which makes it inconvenient for radiologists to distinguish the pathology of the lesions or identify multiple lesions within a single image.
As shown in \figref{fig:rebuttal}, we adopt a bounding box and the proposed \ourmethod\ as a prompt to detect a single lesion for a fair comparison.
Both bounding box and \ourmethod\ can accurately identify and segment the lesion, thus affirming the effectiveness and applicability of \ourmethod.}

\subsection{Issues and Challenges}
This section discusses some common issues, whose visualization and quantitative results are presented in~\figref{fig:qual} and in~\tabref{tab:comparison_weak_annotations}, respectively.
Of note, pancreatic tumor detection and video polyp segmentation is a challenging track in medical imaging, and its overall accuracy is not high enough.
We observe that existing weak annotation strategies (\eg scribble and ellipse) and our \ourmethod\ still lack sufficient robustness.
As for pancreatic tumor detection, our \ourmethod\ fails to capture the camouflaged boundaries of the tumor area by only using the watershed algorithm that utilizes gradient information.
As shown in~\figref{fig:qual}, our \ourmethod\ fails to capture the whole pancreatic tumor due to insignificant appearance changes.
Besides, the non-gradient area on the boundary indicates that the training might cost more iterations to converge and could lose the semantics information due to the insufficient boundary-related representation in such a hard case.
In terms of other weak annotation strategies, the performance of pancreatic tumor detection is relatively lower than \ourmethod\ as shown in~\tabref{tab:comparison_weak_annotations}.
This might be due to the imbalanced data of the tumor area to the whole CT scan and the insufficient boundary supervision.
The aforementioned drawbacks inspire us to explore a more robust annotation strategy to improve the accuracy of pancreatic tumor detection.

On the other hand, despite the generally satisfactory performance of most annotation strategies in video polyp segmentation, reducing annotation time proves challenging due to the absence of a fixed shape for the lesion in 3D spaces, such as the one found in 3D volumetric data. 
Consequently, annotations must be conducted on a slice-by-slice basis.
We also observe that our \ourmethod\ consistently fails to locate lesion regions that share a similar color to the intestinal wall or are too small to be detected.
Thus, there is a large room for improving the detection ability via camouflaged pattern discovery techniques~\cite{fan2021concealed,ji2022gradient} and object tracking techniques~\cite{ciaparrone2020deep}.
In summary, these challenging cases are common difficulties that other methods face and cause severe performance degradation that deserves further exploration.

\section{Conclusion}
To simplify the annotation of temporal and volumetric medical images, we propose a novel annotation strategy called \ourmethod\ and a weakly supervised framework to exploit these annotations. Our proposed strategy can reduce 87.5\% and 99.2\% annotation efforts for polyp and pancreatic tumor detection, respectively, compared with per-pixel annotations. The experimental results demonstrate that our framework achieves a comparable tumor detection rate to per-pixel annotations and higher rates than alternative weak annotation strategies. More importantly, we show that allocating weak annotations from a larger data population, given a certain annotation budget, improves the model generalizability of minority cases compared to per-pixel annotations from a small dataset. We hope our \ourmethod\ strategy can streamline and accelerate the annotation procedure for tumor detection and localization in various medical modalities.

\section*{Acknowledgments}
This work was supported by the Lustgarten Foundation for Pancreatic Cancer Research and the Patrick J. McGovern Foundation Award.

\section*{Conflicts of Interests}
The authors declared that they have no conflicts of interest in this work.
We declare that we do not have any commercial or associative interest that represents a conflict of interest in connection with the work submitted.

\bibliographystyle{IEEEtran}
\bibliography{mir-article}

\begin{thebibliography}{10}
\providecommand{\url}[1]{#1}
\csname url@samestyle\endcsname
\providecommand{\newblock}{\relax}
\providecommand{\bibinfo}[2]{#2}
\providecommand{\BIBentrySTDinterwordspacing}{\spaceskip=0pt\relax}
\providecommand{\BIBentryALTinterwordstretchfactor}{4}
\providecommand{\BIBentryALTinterwordspacing}{\spaceskip=\fontdimen2\font plus
\BIBentryALTinterwordstretchfactor\fontdimen3\font minus
  \fontdimen4\font\relax}
\providecommand{\BIBforeignlanguage}[2]{{%
\expandafter\ifx\csname l@#1\endcsname\relax
\typeout{** WARNING: IEEEtran.bst: No hyphenation pattern has been}%
\typeout{** loaded for the language `#1'. Using the pattern for}%
\typeout{** the default language instead.}%
\else
\language=\csname l@#1\endcsname
\fi
#2}}
\providecommand{\BIBdecl}{\relax}
\BIBdecl

\bibitem{xia2022felix}
Y.~Xia, Q.~Yu, L.~Chu, S.~Kawamoto, S.~Park, F.~Liu, J.~Chen, Z.~Zhu, B.~Li,
  Z.~Zhou \emph{et~al.}, ``The felix project: Deep networks to detect
  pancreatic neoplasms,'' \emph{medRxiv}, pp. 2022--09, 2022.

\bibitem{winawer1993prevention}
S.~J. Winawer, A.~G. Zauber, M.~N. Ho, M.~J. O'brien, L.~S. Gottlieb, S.~S.
  Sternberg, J.~D. Waye, M.~Schapiro, J.~H. Bond, J.~F. Panish \emph{et~al.},
  ``Prevention of colorectal cancer by colonoscopic polypectomy,'' \emph{New
  England Journal of Medicine}, vol. 329, no.~27, pp. 1977--1981, 1993.

\bibitem{rex2017colorectal}
D.~K. Rex, C.~R. Boland, J.~A. Dominitz, F.~M. Giardiello, D.~A. Johnson,
  T.~Kaltenbach, T.~R. Levin, D.~Lieberman, and D.~J. Robertson, ``Colorectal
  cancer screening: recommendations for physicians and patients from the us
  multi-society task force on colorectal cancer,'' \emph{Gastroenterology},
  vol. 153, no.~1, pp. 307--323, 2017.

\bibitem{vazquez2017benchmark}
D.~V{\'a}zquez, J.~Bernal, F.~J. S{\'a}nchez, G.~Fern{\'a}ndez-Esparrach, A.~M.
  L{\'o}pez, A.~Romero, M.~Drozdzal, and A.~Courville, ``A benchmark for
  endoluminal scene segmentation of colonoscopy images,'' \emph{Journal of
  healthcare engineering}, vol. 2017, 2017.

\bibitem{misawa2021development}
M.~Misawa, S.-e. Kudo, Y.~Mori, K.~Hotta, K.~Ohtsuka, T.~Matsuda, S.~Saito,
  T.~Kudo, T.~Baba, F.~Ishida \emph{et~al.}, ``Development of a computer-aided
  detection system for colonoscopy and a publicly accessible large colonoscopy
  video database (with video),'' \emph{Gastrointestinal endoscopy}, vol.~93,
  no.~4, pp. 960--967, 2021.

\bibitem{ma2021ldpolypvideo}
Y.~Ma, X.~Chen, K.~Cheng, Y.~Li, and B.~Sun, ``Ldpolypvideo benchmark: a
  large-scale colonoscopy video dataset of diverse polyps,'' in
  \emph{{International Conference on Medical Image Computing and Computer
  Assisted Intervention}}.\hskip 1em plus 0.5em minus 0.4em\relax {Strasbourg,
  France}: Springer, 2021, pp. 387--396.

\bibitem{smedsrud2021kvasir}
P.~H. Smedsrud, V.~Thambawita, S.~A. Hicks, H.~Gjestang, O.~O. Nedrejord,
  E.~N{\ae}ss, H.~Borgli, D.~Jha, T.~J.~D. Berstad, S.~L. Eskeland
  \emph{et~al.}, ``Kvasir-capsule, a video capsule endoscopy dataset,''
  \emph{Scientific Data}, vol.~8, no.~1, p. 142, 2021.

\bibitem{ji2022video}
G.-P. Ji, G.~Xiao, Y.-C. Chou, D.-P. Fan, K.~Zhao, G.~Chen, and L.~Van~Gool,
  ``Video polyp segmentation: A deep learning perspective,'' \emph{Machine
  Intelligence Research}, vol.~19, no.~6, pp. 531--549, 2022.

\bibitem{lee2017curated}
R.~S. Lee, F.~Gimenez, A.~Hoogi, K.~K. Miyake, M.~Gorovoy, and D.~L. Rubin, ``A
  curated mammography data set for use in computer-aided detection and
  diagnosis research,'' \emph{Scientific data}, vol.~4, no.~1, pp. 1--9, 2017.

\bibitem{porwal2018indian}
P.~Porwal, S.~Pachade, R.~Kamble, M.~Kokare, G.~Deshmukh, V.~Sahasrabuddhe, and
  F.~Meriaudeau, ``Indian diabetic retinopathy image dataset (idrid): a
  database for diabetic retinopathy screening research,'' \emph{Data}, vol.~3,
  no.~3, p.~25, 2018.

\bibitem{qu2023annotating}
C.~Qu, T.~Zhang, H.~Qiao, J.~Liu, Y.~Tang, A.~Yuille, and Z.~Zhou, ``Annotating
  8,000 abdominal ct volumes for multi-organ segmentation in three weeks,''
  \emph{arXiv preprint arXiv:2305.09666}, 2023.

\bibitem{heller2020international}
N.~Heller, S.~McSweeney, M.~T. Peterson, S.~Peterson, J.~Rickman, B.~Stai,
  R.~Tejpaul, M.~Oestreich, P.~Blake, J.~Rosenberg \emph{et~al.}, ``An
  international challenge to use artificial intelligence to define the
  state-of-the-art in kidney and kidney tumor segmentation in ct imaging.''
  2020.

\bibitem{isensee2021nnu}
F.~Isensee, P.~F. Jaeger, S.~A. Kohl, J.~Petersen, and K.~H. Maier-Hein,
  ``nnu-net: a self-configuring method for deep learning-based biomedical image
  segmentation,'' \emph{Nature Methods}, vol.~18, no.~2, pp. 203--211, 2021.

\bibitem{chen2023towards}
J.~Chen, Y.~Xia, J.~Yao, K.~Yan, J.~Zhang, L.~Lu, F.~Wang, B.~Zhou, M.~Qiu,
  Q.~Yu \emph{et~al.}, ``Towards a single unified model for effective
  detection, segmentation, and diagnosis of eight major cancers using a large
  collection of ct scans,'' \emph{arXiv preprint arXiv:2301.12291}, 2023.

\bibitem{liu2023clip}
J.~Liu, Y.~Zhang, J.-N. Chen, J.~Xiao, Y.~Lu, B.~A~Landman, Y.~Yuan, A.~Yuille,
  Y.~Tang, and Z.~Zhou, ``Clip-driven universal model for organ segmentation
  and tumor detection,'' in \emph{{International conference on computer
  vision}}.\hskip 1em plus 0.5em minus 0.4em\relax {Paris, France}: IEEE,
  October 2023, pp. 21\,152--21\,164.

\bibitem{havaei2017brain}
M.~Havaei, A.~Davy, D.~Warde-Farley, A.~Biard, A.~Courville, Y.~Bengio, C.~Pal,
  P.-M. Jodoin, and H.~Larochelle, ``Brain tumor segmentation with deep neural
  networks,'' \emph{Medical image analysis}, vol.~35, pp. 18--31, 2017.

\bibitem{myronenko20193d}
A.~Myronenko, ``3d mri brain tumor segmentation using autoencoder
  regularization,'' in \emph{Brainlesion: Glioma, Multiple Sclerosis, Stroke
  and Traumatic Brain Injuries: 4th International Workshop, BrainLes 2018, Held
  in Conjunction with MICCAI 2018}.\hskip 1em plus 0.5em minus 0.4em\relax
  Granada, Spain: Springer, 2019, pp. 311--320.

\bibitem{wang2021transbts}
W.~Wang, C.~Chen, M.~Ding, H.~Yu, S.~Zha, and J.~Li, ``Transbts: Multimodal
  brain tumor segmentation using transformer,'' in \emph{{International
  Conference on Medical Image Computing and Computer Assisted
  Intervention}}.\hskip 1em plus 0.5em minus 0.4em\relax {Strasbourg, France}:
  Springer, 2021, pp. 109--119.

\bibitem{jiang2022swinbts}
Y.~Jiang, Y.~Zhang, X.~Lin, J.~Dong, T.~Cheng, and J.~Liang, ``Swinbts: A
  method for 3d multimodal brain tumor segmentation using swin transformer,''
  \emph{Brain Sciences}, vol.~12, no.~6, p. 797, 2022.

\bibitem{jin2018ct}
D.~Jin, Z.~Xu, Y.~Tang, A.~P. Harrison, and D.~J. Mollura, ``Ct-realistic lung
  nodule simulation from 3d conditional generative adversarial networks for
  robust lung segmentation,'' in \emph{{International Conference on Medical
  Image Computing and Computer Assisted Intervention}}.\hskip 1em plus 0.5em
  minus 0.4em\relax {Granada, Spain}: Springer, 2018, pp. 732--740.

\bibitem{fan2020inf}
D.-P. Fan, T.~Zhou, G.-P. Ji, Y.~Zhou, G.~Chen, H.~Fu, J.~Shen, and L.~Shao,
  ``Inf-net: Automatic covid-19 lung infection segmentation from ct images,''
  \emph{IEEE Transactions on Medical Imaging}, vol.~39, no.~8, pp. 2626--2637,
  2020.

\bibitem{wang2017central}
S.~Wang, M.~Zhou, Z.~Liu, Z.~Liu, D.~Gu, Y.~Zang, D.~Dong, O.~Gevaert, and
  J.~Tian, ``Central focused convolutional neural networks: Developing a
  data-driven model for lung nodule segmentation,'' \emph{Medical image
  analysis}, vol.~40, pp. 172--183, 2017.

\bibitem{zhu2019multi}
Z.~Zhu, Y.~Xia, L.~Xie, E.~K. Fishman, and A.~L. Yuille, ``Multi-scale
  coarse-to-fine segmentation for screening pancreatic ductal adenocarcinoma,''
  in \emph{{International Conference on Medical Image Computing and Computer
  Assisted Intervention}}.\hskip 1em plus 0.5em minus 0.4em\relax {Shenzhen,
  China}: Springer, 2019, pp. 3--12.

\bibitem{zhang2020robust}
L.~Zhang, Y.~Shi, J.~Yao, Y.~Bian, K.~Cao, D.~Jin, J.~Xiao, and L.~Lu, ``Robust
  pancreatic ductal adenocarcinoma segmentation with multi-institutional
  multi-phase partially-annotated ct scans,'' in \emph{{International
  Conference on Medical Image Computing and Computer Assisted
  Intervention}}.\hskip 1em plus 0.5em minus 0.4em\relax {Lima, Peru}:
  Springer, 2020, pp. 491--500.

\bibitem{li2023early}
B.~Li, Y.-C. Chou, S.~Sun, H.~Qiao, A.~Yuille, and Z.~Zhou, ``Early detection
  and localization of pancreatic cancer by label-free tumor synthesis,''
  \emph{arXiv preprint arXiv:2308.03008}, 2023.

\bibitem{almotairi2020liver}
S.~Almotairi, G.~Kareem, M.~Aouf, B.~Almutairi, and M.~A.-M. Salem, ``Liver
  tumor segmentation in ct scans using modified segnet,'' \emph{Sensors},
  vol.~20, no.~5, p. 1516, 2020.

\bibitem{li2015automatic}
W.~Li \emph{et~al.}, ``Automatic segmentation of liver tumor in ct images with
  deep convolutional neural networks,'' \emph{Journal of Computer and
  Communications}, vol.~3, no.~11, p. 146, 2015.

\bibitem{ji2021progressively}
G.-P. Ji, Y.-C. Chou, D.-P. Fan, G.~Chen, H.~Fu, D.~Jha, and L.~Shao,
  ``Progressively normalized self-attention network for video polyp
  segmentation,'' in \emph{{International Conference on Medical Image Computing
  and Computer Assisted Intervention}}.\hskip 1em plus 0.5em minus 0.4em\relax
  {Strasbourg, France}: Springer, 2021, pp. 142--152.

\bibitem{fan2020pranet}
D.-P. Fan, G.-P. Ji, T.~Zhou, G.~Chen, H.~Fu, J.~Shen, and L.~Shao, ``Pranet:
  Parallel reverse attention network for polyp segmentation,'' in
  \emph{{International Conference on Medical Image Computing and Computer
  Assisted Intervention}}.\hskip 1em plus 0.5em minus 0.4em\relax {Lima, Peru}:
  Springer, 2020, pp. 263--273.

\bibitem{ji2023deep}
G.-P. Ji, D.-P. Fan, Y.-C. Chou, D.~Dai, A.~Liniger, and L.~Van~Gool, ``Deep
  gradient learning for efficient camouflaged object detection,'' \emph{Machine
  Intelligence Research}, vol.~20, no.~1, pp. 92--108, 2023.

\bibitem{wei2021shallow}
J.~Wei, Y.~Hu, R.~Zhang, Z.~Li, S.~K. Zhou, and S.~Cui, ``Shallow attention
  network for polyp segmentation,'' in \emph{{International Conference on
  Medical Image Computing and Computer Assisted Intervention}}.\hskip 1em plus
  0.5em minus 0.4em\relax {Strasbourg, France}: Springer, 2021, pp. 699--708.

\bibitem{papandreou2015weakly}
G.~Papandreou, L.-C. Chen, K.~P. Murphy, and A.~L. Yuille, ``Weakly-and
  semi-supervised learning of a deep convolutional network for semantic image
  segmentation,'' in \emph{{International conference on computer vision}},
  {Santiago, Chile}, 2015, pp. 1742--1750.

\bibitem{hsu2019weakly}
C.-C. Hsu, K.-J. Hsu, C.-C. Tsai, Y.-Y. Lin, and Y.-Y. Chuang, ``Weakly
  supervised instance segmentation using the bounding box tightness prior,''
  \emph{{Advances in neural information processing systems}}, vol.~32, 2019.

\bibitem{zeng2019multi}
Y.~Zeng, Y.~Zhuge, H.~Lu, L.~Zhang, M.~Qian, and Y.~Yu, ``Multi-source weak
  supervision for saliency detection,'' in \emph{{Conference on computer vision
  and pattern recognition}}.\hskip 1em plus 0.5em minus 0.4em\relax {Long
  Beach, CA, USA}: IEEE, 2019, pp. 6074--6083.

\bibitem{wang2017learning}
L.~Wang, H.~Lu, Y.~Wang, M.~Feng, D.~Wang, B.~Yin, and X.~Ruan, ``Learning to
  detect salient objects with image-level supervision,'' in \emph{{Conference
  on computer vision and pattern recognition}}.\hskip 1em plus 0.5em minus
  0.4em\relax {Honolulu, HI, USA}: IEEE, 2017, pp. 136--145.

\bibitem{zhang2020weakly}
J.~Zhang, X.~Yu, A.~Li, P.~Song, B.~Liu, and Y.~Dai, ``Weakly-supervised
  salient object detection via scribble annotations,'' in \emph{{Conference on
  computer vision and pattern recognition}}.\hskip 1em plus 0.5em minus
  0.4em\relax {[Online]]}: IEEE, 2020, pp. 12\,546--12\,555.

\bibitem{chu2021improving}
T.~Chu, X.~Li, H.~V. Vo, R.~M. Summers, and E.~Sizikova, ``Improving weakly
  supervised lesion segmentation using multi-task learning,'' in
  \emph{Proceedings of the Fourth Conference on Medical Imaging with Deep
  Learning}.\hskip 1em plus 0.5em minus 0.4em\relax Lübeck, Germany: PMLR,
  2021, pp. 60--73.

\bibitem{cheng2022pointly}
B.~Cheng, O.~Parkhi, and A.~Kirillov, ``Pointly-supervised instance
  segmentation,'' in \emph{{Conference on computer vision and pattern
  recognition}}.\hskip 1em plus 0.5em minus 0.4em\relax {New Orleans, LA, USA}:
  IEEE, 2022, pp. 2617--2626.

\bibitem{li2019weakly}
C.~Li, X.~Wang, W.~Liu, L.~J. Latecki, B.~Wang, and J.~Huang, ``Weakly
  supervised mitosis detection in breast histopathology images using concentric
  loss,'' \emph{Medical image analysis}, vol.~53, pp. 165--178, 2019.

\bibitem{li2022box2mask}
W.~Li, W.~Liu, J.~Zhu, M.~Cui, R.~Yu, X.~Hua, and L.~Zhang, ``Box2mask:
  Box-supervised instance segmentation via level-set evolution,'' \emph{arXiv
  preprint arXiv:2212.01579}, 2022.

\bibitem{tang2018regularized}
M.~Tang, F.~Perazzi, A.~Djelouah, I.~Ben~Ayed, C.~Schroers, and Y.~Boykov, ``On
  regularized losses for weakly-supervised cnn segmentation,'' in
  \emph{{European conference on computer vision}}.\hskip 1em plus 0.5em minus
  0.4em\relax {Munich, Germany}: Springer, 2018, pp. 507--522.

\bibitem{tang2018normalized}
M.~Tang, A.~Djelouah, F.~Perazzi, Y.~Boykov, and C.~Schroers, ``Normalized cut
  loss for weakly-supervised cnn segmentation,'' in \emph{{Conference on
  computer vision and pattern recognition}}.\hskip 1em plus 0.5em minus
  0.4em\relax {Salt Lake City, UT, USA}: IEEE, 2018, pp. 1818--1827.

\bibitem{chen2022c}
Z.~Chen, Z.~Tian, J.~Zhu, C.~Li, and S.~Du, ``C-cam: Causal cam for weakly
  supervised semantic segmentation on medical image,'' in \emph{{Conference on
  computer vision and pattern recognition}}.\hskip 1em plus 0.5em minus
  0.4em\relax {New Orleans, LA, USA}: IEEE, 2022, pp. 11\,676--11\,685.

\bibitem{liu2022weakly}
X.~Liu, Q.~Yuan, Y.~Gao, K.~He, S.~Wang, X.~Tang, J.~Tang, and D.~Shen,
  ``Weakly supervised segmentation of covid19 infection with scribble
  annotation on ct images,'' \emph{Pattern recognition}, vol. 122, p. 108341,
  2022.

\bibitem{zhang2022cyclemix}
K.~Zhang and X.~Zhuang, ``Cyclemix: A holistic strategy for medical image
  segmentation from scribble supervision,'' in \emph{{Conference on computer
  vision and pattern recognition}}.\hskip 1em plus 0.5em minus 0.4em\relax {New
  Orleans, LA, USA}: IEEE, 2022, pp. 11\,656--11\,665.

\bibitem{lu2020geometry}
W.~Lu, X.~Jia, W.~Xie, L.~Shen, Y.~Zhou, and J.~Duan, ``Geometry constrained
  weakly supervised object localization,'' in \emph{{European conference on
  computer vision}}.\hskip 1em plus 0.5em minus 0.4em\relax Glasgow, UK:
  Springer, 2020, pp. 481--496.

\bibitem{tian2021boxinst}
Z.~Tian, C.~Shen, X.~Wang, and H.~Chen, ``Boxinst: High-performance instance
  segmentation with box annotations,'' in \emph{{Conference on computer vision
  and pattern recognition}}.\hskip 1em plus 0.5em minus 0.4em\relax
  {[Online]]}: IEEE, 2021, pp. 5443--5452.

\bibitem{roth2021going}
H.~R. Roth, D.~Yang, Z.~Xu, X.~Wang, and D.~Xu, ``Going to extremes: weakly
  supervised medical image segmentation,'' \emph{Machine Learning and Knowledge
  Extraction}, vol.~3, no.~2, pp. 507--524, 2021.

\bibitem{zhou2019integrating}
Z.~Zhou, J.~Shin, R.~Feng, R.~T. Hurst, C.~B. Kendall, and J.~Liang,
  ``Integrating active learning and transfer learning for carotid intima-media
  thickness video interpretation,'' \emph{Journal of digital imaging}, vol.~32,
  no.~2, pp. 290--299, 2019.

\bibitem{fitzgibbon1996buyer}
A.~W. Fitzgibbon, R.~B. Fisher \emph{et~al.}, \emph{A buyer's guide to conic
  fitting}.\hskip 1em plus 0.5em minus 0.4em\relax Citeseer, 1996.

\bibitem{meyer1994topographic}
F.~Meyer, ``Topographic distance and watershed lines,'' \emph{Signal
  processing}, vol.~38, no.~1, pp. 113--125, 1994.

\bibitem{kirillov2023segment}
A.~Kirillov, E.~Mintun, N.~Ravi, H.~Mao, C.~Rolland, L.~Gustafson, T.~Xiao,
  S.~Whitehead, A.~C. Berg, W.-Y. Lo \emph{et~al.}, ``Segment anything,''
  \emph{arXiv preprint arXiv:2304.02643}, 2023.

\bibitem{ma2023segment}
J.~Ma and B.~Wang, ``Segment anything in medical images,'' \emph{arXiv preprint
  arXiv:2304.12306}, 2023.

\bibitem{fan2021concealed}
D.-P. Fan, G.-P. Ji, M.-M. Cheng, and L.~Shao, ``Concealed object detection,''
  \emph{IEEE transactions on pattern analysis and machine intelligence},
  vol.~44, no.~10, pp. 6024--6042, 2021.

\bibitem{ji2022gradient}
G.-P. Ji, D.-P. Fan, Y.-C. Chou, D.~Dai, A.~Liniger, and L.~Van~Gool, ``Deep
  gradient learning for efficient camouflaged object detection,'' \emph{Machine
  Intelligence Research}, vol.~20, no.~1, pp. 92--108, 2023.

\bibitem{ciaparrone2020deep}
G.~Ciaparrone, F.~L. S{\'a}nchez, S.~Tabik, L.~Troiano, R.~Tagliaferri, and
  F.~Herrera, ``Deep learning in video multi-object tracking: A survey,''
  \emph{Neurocomputing}, vol. 381, pp. 61--88, 2020.

\end{thebibliography}

\vspace{20pt}

\vspace{20pt}
\noindent{\bf Yu-Cheng Chou} 
is currently a Ph.D. student at Johns Hopkins University, supervised by Zongwei Zhou and Prof. Alan Yuille.
He received a Bachelor's degree in Software Engineering at the School of Computer Science, Wuhan University, in 2022.
Drawing upon medical imaging, causality, and computer vision, his research focuses on developing novel methodologies to accurately detect lesions and exploring explainability through causality for computer-aided diagnosis and surgery.

E-mail: \href{johnson111788@gmail.com}{johnson111788@gmail.com}

ORCID iD: 0000-0002-9334-2899

\vspace{20pt}
\noindent{\bf Bowen Li} is currently a Ph.D. student at Johns Hopkins University, supervised by Zongwei Zhou and Prof. Alan Tuille. He received a Bachelor's degree in Automation at Tsinghua University and a Master's degree in Computer Science at Johns Hopkins University. His research focuses on computer-aided diagnosis with ultrasound and CT.

E-mail: \href{lbwdruid@gmail.com}{lbwdruid@gmail.com}

ORCID iD: 0000-0003-1374-5207

\vspace{20pt}
\noindent{\bf Deng-Ping Fan} 
received his PhD degree from Nankai University in 2019.
He joined the Inception Institute of Artificial Intelligence (IIAI) in 2019.
He has published about 50 top journal and conference papers such as TPAMI, IJCV, TIP, TNNLS, TMI, CVPR, ICCV, ECCV, IJCAI, etc.
His research interests include computer vision, deep learning, and saliency detection.
He won the Best Paper Finalist Award at IEEE CVPR 2019, and the Best Paper Award Nominee at IEEE CVPR 2020.
He was recognized as the CVPR 2019 outstanding reviewer with a special mention award, the CVPR 2020 outstanding reviewer, the ECCV 2020 high-quality reviewer, and the CVPR 2021 outstanding reviewer.
He served as a program committee board (PCB) member of IJCAI 2022-2024, a senior program committee (SPC) member of IJCAI 2021, a program committee member (PC) of CAD\&CG 2021, a committee member of China Society of Image and Graphics (CSIG), area chair in NeurIPS 2021 Datasets and Benchmarks Track, area chair in MICCAI2020 Workshop.

E-mail: \href{dengpfan@gmail.com}{dengpfan@gmail.com}

ORCID iD: 0000-0002-5245-7518

\vspace{20pt}
\noindent{\bf Alan Yuille} received the BA degree in mathematics from the University of Cambridge in 1976. His PhD on theoretical physics, supervised by Prof. S.W. Hawking, was approved in 1981. He was a research scientist in the Artificial Intelligence Laboratory at MIT and the Division of Applied Sciences at Harvard University from 1982 to 1988. He served as an assistant and associate professor at Harvard until 1996. He was a senior research scientist at the Smith-Kettlewell Eye Research Institute from 1996 to 2002. He was a full professor of Statistics at the University of California, Los Angeles, as a full professor with joint appointments in computer science, psychiatry, and psychology. He moved to Johns Hopkins University in January 2016. His research interests include computational models of vision, mathematical models of cognition, medical image analysis, and artificial intelligence and neural networks.

E-mail: \href{ayuille1@jhu.edu}{ayuille1@jhu.edu}

ORCID iD: 0000-0001-5207-9249

\vspace{20pt}
\noindent{\bf Zongwei Zhou} 
is a postdoctoral researcher at Johns Hopkins University. He received his Ph.D. in Biomedical Informatics at Arizona State University in 2021. His research focuses on developing novel methods to reduce the annotation efforts for computer-aided detection and diagnosis. Zongwei received the AMIA Doctoral Dissertation Award in 2022, the Elsevier-MedIA Best Paper Award in 2020, and the MICCAI Young Scientist Award in 2019. In addition to five U.S. patents, Zongwei has published over thirty peer-reviewed journal/conference articles, two of which have been ranked among the most popular articles in IEEE TMI and the highest-cited article in EJNMMI Research. He was named the top 2\% of Scientists released by Stanford University in 2022 and 2023. 

E-mail: \href{zzhou82@jh.edu}{zzhou82@jh.edu}

ORCID iD: 0000-0002-3154-9851

\end{document}